# Convolutional Fourier Analysis Network (CFAN): A Unified Time-Frequency Approach for ECG Classification

**Sam Jeong[1], and Hae Yong Kim[2]**

[1,2]Department of Electronic Systems Engineering, Polytechnic School, University of São Paulo, Brazil

Corresponding author: Sam Jeong (e-mail: sam.jeong@usp.br).

This work was financed in part by the Coordination for the Improvement of Higher Education Personnel - Brazil (CAPES) - Finance Code 001 and through PhD scholarship number 88887.838214/2023-00. The Article Processing Fee for the publication of this research was covered by the Coordination for the Improvement of Higher Education Personnel - CAPES (ROR identifier: 00x0ma614).

**ABSTRACT** Machine learning has revolutionized biomedical signal analysis, particularly in electrocardiogram (ECG) classification. While convolutional neural networks (CNNs) excel at automatic feature extraction, the optimal integration of time- and frequency-domain information remains unresolved. This study introduces the Convolutional Fourier Analysis Network (CFAN), a novel architecture that unifies time-frequency analysis by embedding Fourier principles directly into CNN layers. We evaluate CFAN against four benchmarks — spectrogram-based 2D CNN (SPECT); 1D CNN (CNN1D); Fourier-based 1D CNN (FFT1D); and CNN1D with integrated Fourier Analysis Network (CNN1D-FAN) — across three ECG tasks: arrhythmia classification (MIT-BIH), identity recognition (ECG-ID), and apnea detection (Apnea-ECG). CFAN achieved state-of-the-art performance, surpassing all competing methods with accuracies of 98.95% (MIT-BIH), 96.83% (ECG-ID), and 95.01% (Apnea-ECG). Notably, on ECG-ID and Apnea-ECG, CFAN demonstrated statistically significant improvements over the second-best method (CNN1D-FAN, $p \leq 0.02$), further validating its superior performance. Key innovations include CONV-FAN blocks that combine *sine*, *cosine* and GELU activations in convolutional layers to capture periodic features and joint time-frequency learning without spectrogram conversion. Our results highlight CFAN's potential for broader biomedical and signal classification applications.

**INDEX TERMS** Apnea detection, arrhythmia classification, convolutional neural network, deep learning, electrocardiogram, Fourier Analysis Network, identity recognition, machine learning, time-frequency analysis.

## I. INTRODUCTION

Machine learning has become increasingly prevalent in medical applications, particularly for biomedical signal analysis. In classifying biomedical signals, features are typically extracted from the time domain, frequency domain, or their combination, which are then used as inputs for various classification algorithms. Early approaches predominantly employed conventional machine learning classifiers, including linear discriminant analysis [1]–[3], decision trees [1], k-nearest neighbors [1], and support vector machines [1], [2], [4], [5].

The advent of deep learning, particularly convolutional neural networks (CNNs) [6]-[27], marked a significant shift in biomedical signal classification. Unlike traditional classifiers, CNNs automatically extract and select relevant features from input data, minimizing reliance on manual feature engineering. Initially, CNNs were designed for image-based tasks, requiring biomedical signals to be converted into image-like representations, such as spectrograms, using techniques like short-time Fourier transform (STFT) or wavelet transform [6]-[11]. However, the development of one-dimensional CNNs [12]-[23] has enabled the direct analysis of raw time-series signals, eliminating the need for prior transformation.

Consequently, CNNs can classify biomedical signals across three distinct domains: (a) time-domain classification using a 1D CNN (CNN1D); (b) frequency-domain classification with a Fourier transform-based 1D CNN (FFT1D); and (c) time-frequency classification using a spectrogram-based 2D CNN (SPECT).

The electrocardiogram (ECG) provides a non-invasive measurement of cardiac electrical activity and has served as the clinical gold standard for arrhythmia detection and classification [1], [2], [4], [6]-[10], [12]-[15]. Recent advances have extended its applications to diverse domains, including: biometric identification [3], [5], [16]-[18], [24], [25], sleep stage monitoring [11], [19]-[23], [26], [27], stress quantification [28], and affective computing through emotion recognition [29]. In this work, we test CNNs across three clinically relevant tasks using PhysioNet ECG datasets [30]: (i) arrhythmia classification (MIT-BIH [31]), (ii) subject identification (ECG-ID [32]), and (iii) sleep apnea detection (Apnea-ECG [33]).

We rigorously optimized our approach to achieve state-of-the-art classification performance across all tasks and domains. Contrary to our initial hypothesis that SPECT (time-frequency domain) would perform best, CNN1D (time-domain only) achieved superior ECG classification accuracy. This suggests spectrograms are suboptimal for integrating time-frequency information. This result demonstrates the need for a more effective strategy to integrate time and frequency domains.

To accomplish this objective, we integrated the Fourier Analysis Network (FAN) [34] – a recently proposed architecture currently available as a preprint – by systematically replacing all fully-connected layers in our baseline CNN1D model with fully-connected FAN (FC-FAN) blocks, obtaining CNN1D-FAN model. This architectural modification enables simultaneous learning in both time and frequency domains through the network's inherent Fourier analysis capabilities. CNN1D-FAN demonstrated only marginal improvement over CNN1D, with less than 0.1% difference in mean accuracy across all evaluated tasks.

This marginal improvement motivated the development of our Convolutional Fourier Analysis Network (CFAN), a novel architecture that synergistically combines Fourier Analysis Networks (FAN) with convolutional neural networks (CNNs) through specialized CONV-FAN blocks. The accuracy of CFAN outperformed all other techniques across all evaluated tasks, with a statistically significant improvement ($p \leq 0.02$) over the second-best method (CNN1D-FAN) in two out of the three problems. In the third problem, accuracy was near the ceiling (close to 100%), which prevents meaningful hypothesis testing. CFAN achieved state-of-the-art performance across all tasks, conclusively validating the benefits of joint time-frequency feature learning. We hypothesize that CFAN's architecture has broader applicability beyond ECG analysis, potentially extending to various signal processing domains where time-frequency representations are crucial.

Since our CFAN implementation differs slightly from the original FAN architecture proposed in [34], we also evaluated the original configuration (which we term CSFAN - Convolutional Shared-matrix FAN). Our experimental results demonstrate that CFAN outperforms its variant CSFAN, validating our architectural modifications.

The remainder of this paper is organized as follows. Section II reviews: prior works analyzing biomedical signals in both time and frequency domains; the recently proposed Fourier Analysis Network (FAN); and the three ECG datasets used in this study along with comparative studies employing these datasets. Section III describes the five benchmark methods compared against our proposed CFAN. Section IV presents experimental results and statistical validation of CFAN's superior performance. Finally, section V provides conclusions.

## II. RELATED WORKS AND ECG DATASETS

### A. CURRENT TIME-FREQUENCY ECG ANALYSIS

Time-frequency analysis of ECG signals has become fundamental for biomedical classification tasks. Existing methodologies fall into three principal categories:

1) Feature-based approaches: Traditional machine learning techniques employ handcrafted feature extraction, as demonstrated by Elhaj *et al.* [4], who combined temporal and spectral features to train SVM classifiers for heartbeat classification.
2) Image-based representations: Deep learning methods frequently transform 1D signals into 2D representations. Demčáková *et al.* [6] make use of this approach by converting ECGs into time-series plots and STFT spectrograms for CNN processing, effectively translating the classification problem into computer vision domains.
3) Hybrid architectures: Recent advances have introduced sophisticated hybrid approaches that combine the strengths of multiple methodologies. (a) Kumari and Goswami [8] used CNN-KNN to analyze superlet-transform spectrograms; (b) Sharma *et al.* [9] compared CNN-BiLSTM models against traditional classifiers using engineered features; (c) Chen *et al.* [10] fused VGG networks for spectrogram analysis with LSTMs networks for raw time-series processing, for atrial fibrillation detection.

These studies collectively demonstrate the importance of both temporal and spectral information, though optimal integration strategies remain an open challenge. This gap motivates our development of CFAN, which fundamentally rethinks time-frequency fusion through learnable Fourier analysis blocks.

### B. FOURIER ANALYSIS NETWORK (FAN)

Fourier Analysis Network (FAN) [34] models periodic information using Fourier Series, by incorporating *sine* and *cosine* activation functions in fully connected (FC) layers. The authors designed FAN to satisfy two principles:

1) The capacity of FAN to represent the Fourier coefficients should be positively related to its depth;

2) The output of any hidden layer can be employed to model periodicity using Fourier Series through the subsequent layers.

After presenting several arguments, they ended up defining the FAN layer as (Fig. 1a):

$$\phi_S(x) \equiv \left[\, cos(W_p x) \,||\, sin(W_p x) \,||\, \sigma(B_{\bar{p}} + W_{\bar{p}} x) \,\right] \tag{1}$$

where $W_p \in \mathbb{R}^{d_p \times d_x}$, $W_{\bar{p}} \in \mathbb{R}^{d_{\bar{p}} \times d_x}$ and $B_{\bar{p}} \in \mathbb{R}^{d_{\bar{p}}}$ are learnable parameters (with the hyperparameters $d_p$ and $d_{\bar{p}}$ indicating the first dimension of $W_p$ and $W_{\bar{p}}$, respectively); the input vector $x \in \mathbb{R}^{d_x}$; the output layer $\phi_S(x) \in \mathbb{R}^{2d_p + d_{\bar{p}}}$; $\sigma$ denotes the activation function (the paper adopts GELU — Gaussian Error Linear Unit); and $[\cdot||\cdot]$ denotes the concatenation along the first matrix dimension. To prevent ambiguity with other layer types, we will refer to this as the Fully Connected Shared-matrix FAN (FC-SFAN) layer throughout this paper. The entire original FAN model is defined as the stacking of the FC-SFAN layers.

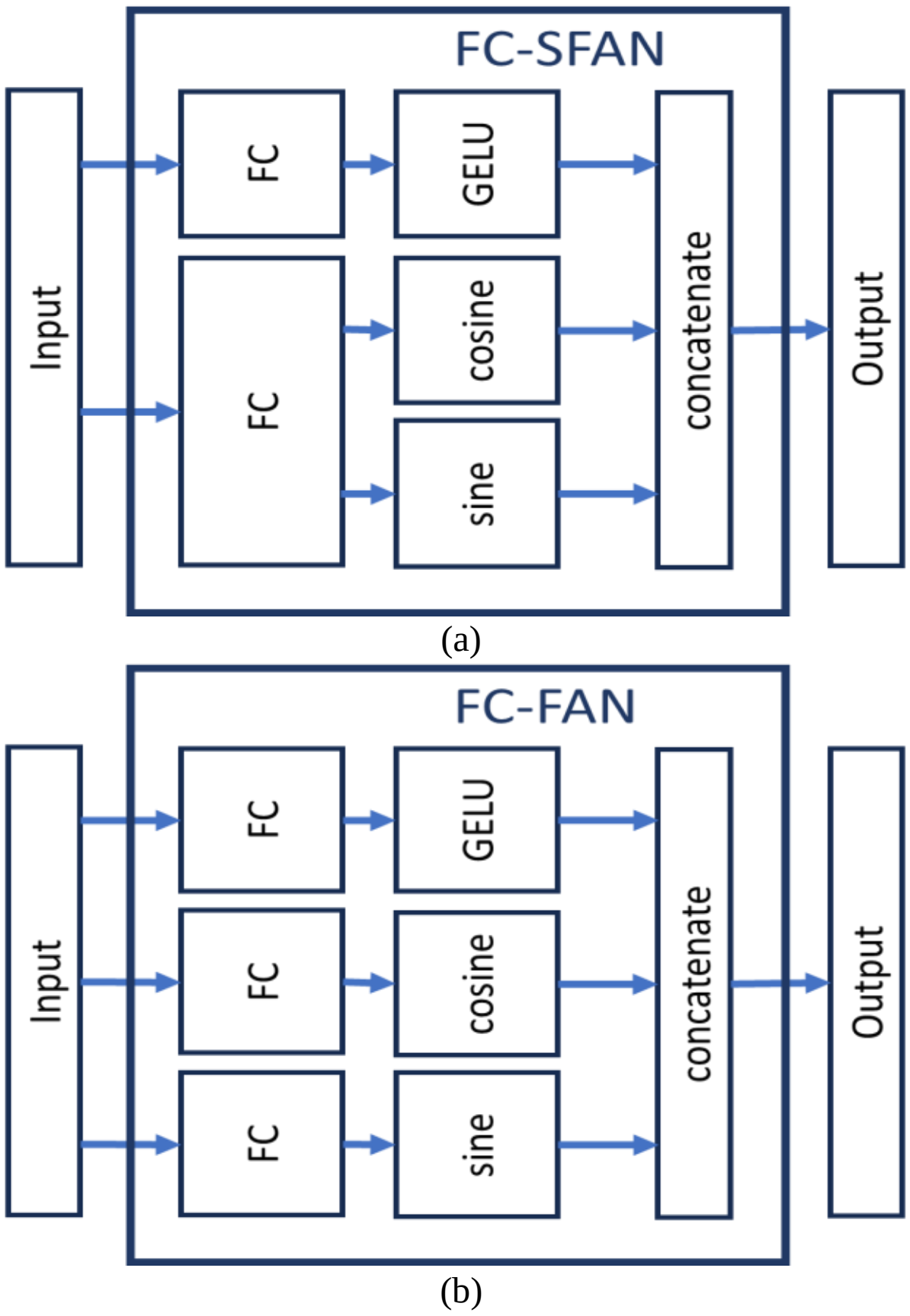

**FIGURE 1. (a) Original FAN layer employs shared *sine*/*cosine* weights with a neuron distribution of 4:1 (GELU to *sine*/*cosine* activations). To avoid ambiguity, we call it FC-SFAN (Fully Connected Shared-matrix FAN) layer. (b) Our proposed FC-FAN layer uses independent *sine* and *cosine* weights, with a neuron ratio of 4:1:1 (GELU to *sine* to *cosine* activations).**

The authors argue that the FC-SFAN layer $\phi_S(x)$ can seamlessly replace a fully connected (FC) layer. So, rather than defining a specific network architecture, FAN modifies existing networks by replacing FC layers with standard activations by a combination of GELU and *sine*/*cosine* activation functions in a 4:1 ratio.

FAN can model periodic signals with fewer parameters than multilayer perceptron with FC layers, while achieving comparable performance [34]. This efficiency stems from its use of *sine* and *cosine* functions, mirroring Fourier analysis' representation of periodic signals. To illustrate this concept, consider the Fourier series representation of a signal *f*(*t*):

$$f(t) = C_0 + \sum_{n=1}^{\infty}\left(a_n \, cos(w_n t) + b_n sin(w_n t)\right), \tag{2}$$

where:

$$C_0 = \frac{1}{T}\int_0^T f(t)\, dt, \tag{3}$$

$$a_n = \frac{2}{T}\int_0^T f(t) cos(w_n t)\, dt, \tag{4}$$

$$b_n = \frac{2}{T}\int_0^T f(t) sin(w_n t)\, dt. \tag{5}$$

Fig. 2 illustrates how Fourier series principles are embodied in FAN's architecture. Here, the first neuron outputs $cos(wt)$ while the second outputs $sin(wt)$. The next layer combines these as $C_0 + a\, cos(wt) + b\, sin(wt)$, a fundamental Fourier series expansion (*n*=1) of the input signal.

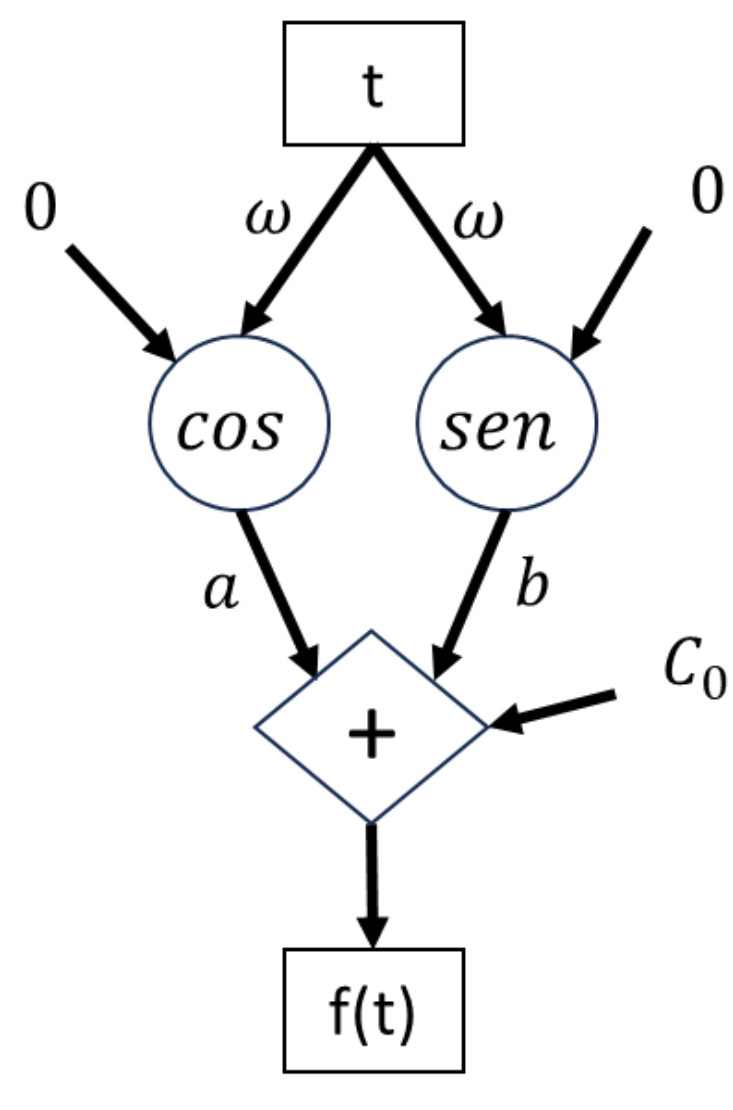

**FIGURE 2. Example illustrating how FAN incorporates Fourier series principles.**

The original FAN paper [34] primarily focuses on modeling periodicity, especially in out-of-domain scenarios. The authors generate periodic functions, train the model within a specific domain, and evaluate its ability to predict behavior beyond that domain. Their results demonstrate that FAN significantly outperforms other architectures on this task, likely because it is particularly well-suited to Fourier-based representations.

The authors also highlight FAN's versatility by applying it to a variety of real-world tasks, including symbolic formula representation, time series forecasting, language modeling, and image recognition. In these applications, the reported performance gains are generally modest. This contrasts with periodicity modeling, where FAN demonstrates a clear and substantial advantage over other techniques. Nonetheless, the improvements in real-world scenarios, while moderate, appear to be consistent and measurable. This aligns closely with our own findings in ECG classification experiments.

Our empirical validation in ECG classification revealed that employing separate weight matrices for *sine* ($W_s$) and *cosine* ($W_c$) components yield performance improvements over using a single matrix for both activations (see Section IV-B). So, throughout the remainder of this work, we employ the following equation with distinct *sine* and *cosine* weight matrices (Fig. 1b), named FC-FAN layer, unless otherwise stated.

$$\phi(x) \equiv \left[\, cos(W_c x) \,||\, sin(W_s x) \,||\, \sigma(B_{\overline{p}} + W_{\overline{p}} x) \,\right] \tag{6}$$

### C. MIT-BIH DATASET

The MIT-BIH Arrhythmia Database [31], published in 1982, is one of the most widely used datasets for arrhythmia detection and classification. It comprises 48 ECG recordings from 47 individuals, sampled at 360 Hz. Each recording lasts approximately 30 minutes and is paired with an annotation file that details heartbeat classifications and the type, start, and end points of heart rhythms determined by at least two cardiologists.

From these signals, we utilized the pre-annotated R-peaks to extract 257-sample segments, with the R-peak centered. Following the Advancement of Medical Instrumentation guidelines [35], each beat segment was categorized into five groups: Normal, Supraventricular ectopic beat, Ventricular ectopic beat, Fusion, and Unknown beat, respectively with 90593, 2781, 7235, 802, and 8040 segments with a total of 109,451. Fig. 3(a) presents an example of a preprocessed ECG segment of this dataset.

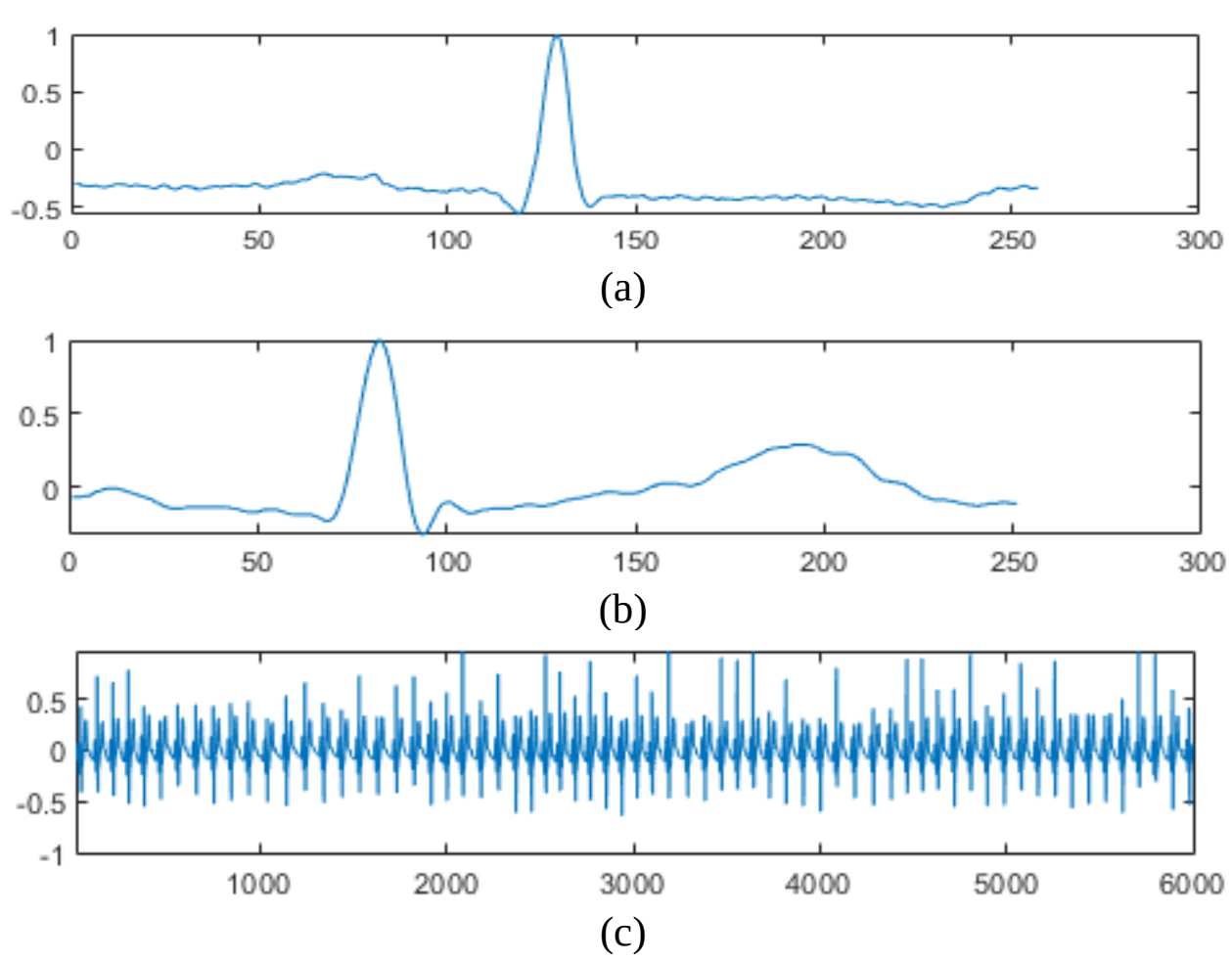

**FIGURE 3. Examples of preprocessed ECG segments, normalized by subtracting the mean value and dividing by the maximum amplitude. (a) MIT-BIH Arrhythmia, (b) ECG-ID, and (c) Apnea-ECG.**

Below, we list several studies that utilize this dataset.

Elhaj *et al*. [4] used segments of 100 samples before to 200 samples after the R-peak to extract handcrafted time- and frequency-domain features as inputs to SVM classifiers to obtain accuracy of 98.95%. Although they used the same group categorizations for classification, the resulting numbers of samples for each class differed significantly from those in our study, making direct comparison difficult.

Demčáková *et al*. [6] used segments of 180 samples before to 216 after the R-peak and generated time-domain images as inputs to CNN to accuracy of 98.50% and average precision, recall, and F1-score of 95.17%, 91.65% and 93.38% respectively. They used 6 categories of beats, thus is not directly comparable with our study.

Sharma *et al*. [9] used segments with size of 500 samples to obtain accuracy of 98.67%, and average precision, recall and F1-score of 97.8%, 96.0% and 96.80% respectively.

Nemissi *et al*. [13] used 160 samples before to 99 after R-peak as inputs to CNN with handcrafted features concatenated with the convolutional features to obtain accuracy of 98.78% and average precision, recall, and F1-score of 92.61%, 94.61% and 93.85% respectively.

Farag [14] achieved 99.08% accuracy and near-perfect average precision, recall, and F1-scores (99.09%, 99.08%, and 99.09% respectively) using a CNN with 64-sample R-peak-centered segments augmented by local and global RR interval features. Their subsequent inter-patient study [15] reported slightly reduced performance (98.18% accuracy, with precision/recall/F1-scores of 92.44%, 91.9%, and 92.17%). These results are not directly comparable to our work, as Farag's approach incorporates additional temporal features (local and global RR intervals) that enhance classification performance. While Farag's inter-patient analysis represents valuable work in this domain, our study follows the predominant literature convention of mixed training/testing sets to ensure broader comparability.

### D. ECG-ID DATASET

The second database we used was the ECG-ID [32] designed for identity recognition. This dataset contains 310 single-lead ECG recordings from 90 individuals, each lasting 20 seconds and sampled at 500 Hz. R-peaks of the QRS complex were detected using the Pan-Tompkins method [36]. Using the same segmentation used by Nemirko and Lugovaya [3], for each cardiac cycle, we extracted a waveform consisting of 80 samples before and 170 samples after each R-peak. For each recording, the eight cycles with the smallest Euclidean distances from the average were selected. Each selected cycle was further processed by subtracting its mean value. This approach resulted in a total of 2,456 cardiac cycles across 90 distinct classes (individuals). Fig. 4 shows the number of samples per class obtained from the ECG-ID database. Fig. 3(b) presents an example of a preprocessed ECG segment of this dataset.

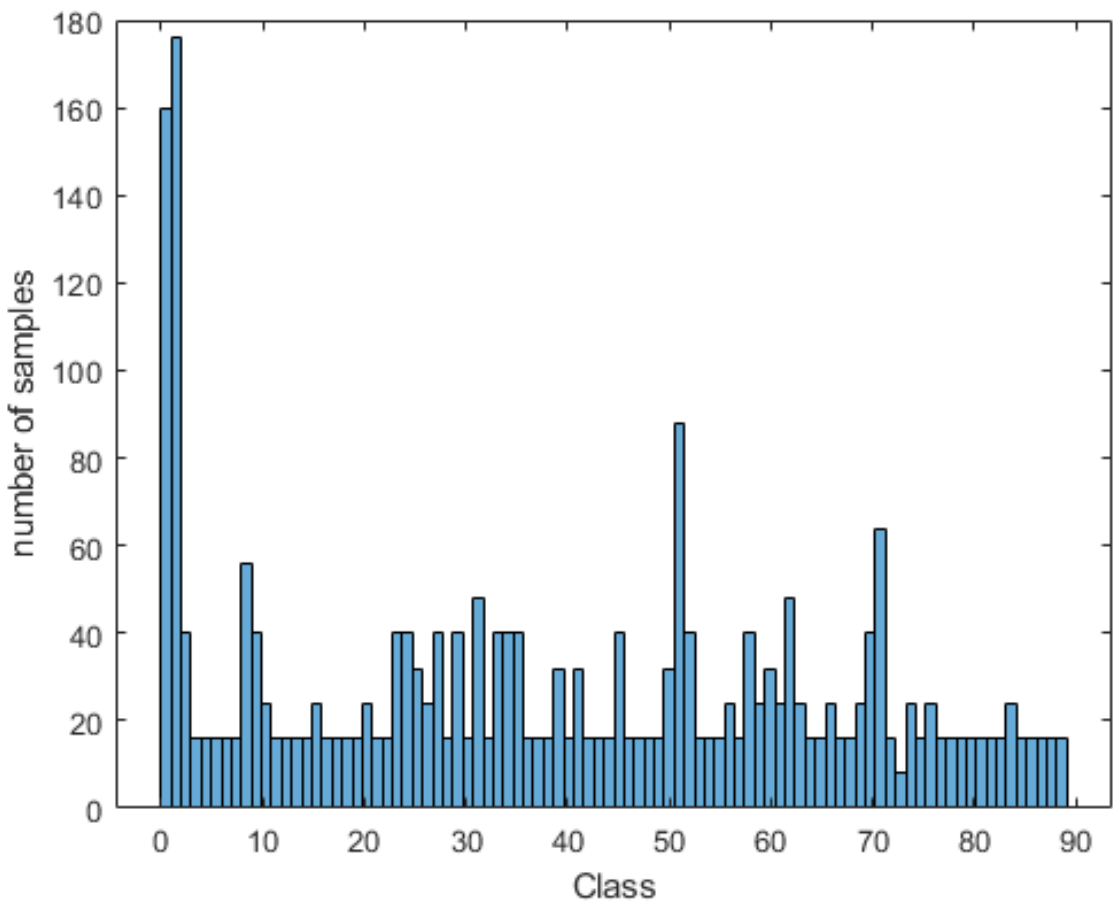

**FIGURE 4.** **Number of samples of each class obtained from the ECG-ID database.**

Previous studies have employed this dataset with varying methodologies.

Nemirko and Lugovaya [3] used features extracted from the segments as inputs to Linear Discriminant Analysis to obtain accuracy of 96%.

Fatima *et al.* [5] systematically evaluated segment length optimization for cardiac cycle analysis, comparing time- and frequency-domain features as Random Forest classifier inputs. Their study defined a cardiac cycle as comprising 125 samples preceding and 200 samples following the R-peak. The optimal configuration utilizing two complete cardiac cycles achieved 97.92% classification accuracy, while single-cycle analysis resulted in significantly reduced performance (93.41%), demonstrating the importance of temporal context in feature extraction. Our study uses a single-cycle of beat to classify, thus being comparable with the last performance.

Zhao *et al.* [16] developed a hybrid deep learning approach employing 5-cycle segments that were either zero-padded or truncated to maintain a fixed length of 1,500 samples. These standardized segments served as input to a convolutional neural network (CNN), which generated embedded representations subsequently classified using k-nearest neighbors (k-NN), achieving a classification accuracy of 97.13%. Zhao *et al.* do not report results using the single-cycle approach adopted in our study, and Fatima *et al.* [5] have demonstrated that utilizing multiple cycles leads to improved performance. Consequently, their results are not directly comparable to ours.

Yi *et al.* [18] implemented a CNN-based biometric identification system using fixed-length ECG segments comprising 119 samples preceding and 180 samples following the R-peak. This approach achieved 94.26% recognition accuracy.

### *E. APNEA-ECG DATASET*

The third database, Apnea-ECG, is one of the most widely used datasets for apnea detection [33]. It comprises 70 ECG recordings sampled at 100 Hz, divided into training and test groups. For this study, we used 35 recordings from the training group. Each recording is paired with a file containing expert annotations indicating the presence or absence of apnea for each minute, based on concurrently recorded breathing and related signals.

To preprocess the ECG segments, we first applied a third-order Savitzky-Golay filter with a window length of five, following the method described by Misra *et al.* [26] and Bhongade *et al.* [19]. This filter effectively removes high-frequency noise while preserving low-frequency components. Finally, each segment was normalized by subtracting its mean and dividing by its standard deviation.

We segmented each training recording into 1-minute intervals, excluding those with signal losses exceeding 0.5 seconds. This process resulted in 15,880 one-minute segments (6,000 sample points each), with 5,925 labeled as apnea and 9,955 as normal. Fig. 3(c) presents an example of a preprocessed ECG segment of this dataset.

Despite using the same dataset and segment size, prior studies have reported varying numbers of extracted segments. Below, we list several studies that utilize this dataset.

Zhao *et al.* [22] developed a CNN architecture with temporal-wise attention mechanisms for sleep apnea detection, utilizing 5-minute signal segments (30,000 samples each). Their approach incorporated both RR intervals and R-peak amplitudes as input features, achieving 91.02% accuracy, 84.25% recall, 95.09% specificity, and 87.56% F1-score. The different size of segments used as input (5 min vs 1 min) of the classifier make these results not directly comparable to ours.

Nguyen *et al.* [23] used 5 segments of 1-minute intervals (6,000 samples each) to generate a matrix of distances of R-peaks used as input to CNN to obtain 92.1% accuracy, 87.7% recall, and 94.8% specificity. Nguyen *et al.* didn't use information contained in segments of 1-minute, but incorporated information of 4 adjacent segments, as input to classifier and used training and test set to obtain a total of 33,659 segments, making these results not directly comparable to ours.

Zhou *et al.* [11] used segments of 1-minute intervals transformed to scalogram images with continuous wavelet transform and to Gramian Angular Field matrix image. Both were used as inputs to different CNNs and a voting system was used to obtain the final result. They obtained 90.93% accuracy, 83.68% recall, and 95.29% specificity. Although Zhou *et al.* used segments of 1-minute intervals, they combined the training and testing groups as Nguyen *et al.* [23] to obtain a slightly higher number of segments (34103), thus these results are not directly comparable to ours.

Misra *et al.* [26] and Bhongade *et al.* [19] used segments of 1-minute intervals (6,000 samples each) directly as inputs to CNN, using different CNN architectures to detect apnea. Misra *et al*. obtained 93.01% accuracy, 92.93% recall and 93.10% specificity. Bhongade *et al*. obtained 94.77% accuracy, 92.55% recall, 93.80% precision, 96.14% specificity and 93.07% F1-score. Even using the training and test groups as Nguyen *et al*. [23] and Zhou *et al*. [11] Misra *et al*. obtained a total of 11,620 segments. However, Bhongade *et al*. used only the training group and segments of 1-minute intervals, thus the results presented by them are comparable to our study.

Li *et al*. [20] used segments of 1-minute intervals to extract features to use as inputs to dynamical estimators consisting of neural networks to obtain 98.31% accuracy, 97.66% recall, 98.56% specificity, and 96.45% precision. As Misra *et al*. [26], Li *et al*. combined the training and test group to produce a new set of 11,690 segments - a number that differs substantially from the segment count used in our work. Therefore, their results are not directly comparable to ours.

### F. K-FOLD CROSS VALIDATION

For MIT-BIH and Apnea-ECG, the data was split into 10 stratified folds. In each training iteration, 1 fold was reserved for validation and testing (evenly split between the two), while the remaining 9 folds were used for training. Due to the smaller sample size of the ECG-ID dataset, it was divided into 4 stratified folds, following the same training and testing procedure.

## III. PROPOSED METHOD

### A. CLASSIFICATION METHODS

We evaluated five distinct approaches to address the three ECG classification challenges:

1) SPECT: A spectrogram-based 2D CNN classifier operating in the frequency-time domain.
2) CNN1D: A 1D CNN for direct time-domain signal classification.
3) FFT1D: The CNN1D architecture applied to frequency-domain representation.
4) CNN1D-FAN: A modified CNN1D architecture where fully connected (FC) layers were replaced with FC-FAN layers with distinct *sine* and *cosine* matrices.
5) CFAN (Convolutional FAN): This architecture replaces convolutional layers in CNN1D-FAN model with CONV-FAN layers. In this implementation, the *sine* and *cosine* activation functions within each block (both fully connected and convolutional) uses independent weight matrices.

CFAN demonstrates superior classification performance, consistently outperforming all competing methods across key metrics.

We also evaluate a variant of CFAN named CSFAN (Convolutional Shared-matrix FAN). This variant replaces both FC-FAN and CONV-FAN blocks in CFAN with their shared-weight counterparts (FC-SFAN and CONV-SFAN), where a single weight matrix is jointly used for both sine and cosine activations within each block.

### B. 2-D SPECTROGRAM CLASSIFICATION (SPECT)

Convolutional Neural Networks (CNNs) were originally developed for image classification. Early applications of CNNs to biomedical signal classification involved converting signals into spectrogram images to serve as network inputs [37]. The Short-Time Fourier Transform (STFT) applies the FFT over a moving window to generate a time-dependent frequency representation. A spectrogram visualizes the STFT by representing the magnitude of the frequency spectrum for each window as a vertical line in an image. This results in a combined time-frequency representation, where the horizontal axis corresponds to time, the vertical axis to frequency, and the color or intensity to magnitude.

In this study, we used the implementation from the Python SciPy library to generate STFT for the Tensorflow environment, applying a Hann window of size 64 with an overlap of 48. The resulting spectrograms were resized to 64×64 across all datasets. To optimize memory usage during training and testing, spectrograms were generated in grayscale (Fig. 5).

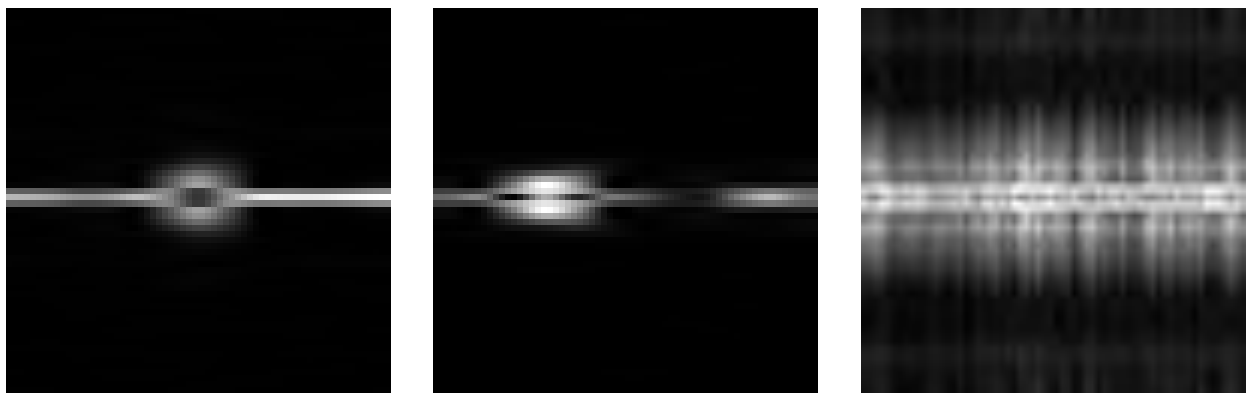

**FIGURE 5.** Number of samples of each class obtained from the ECG-ID database.

To classify the spectrograms, we applied transfer learning using EfficientNetB0, a model pretrained on the ImageNet database with natural images. We replaced top EfficientNetB0 layers with a fully connected (FC) layer using ReLU activation, followed by another FC layer with Softmax activation. We chose the architecture and parameter that achieved the highest classification performance after extensive experimentation with various models.

### C. 1-D TIME SEQUENCE CLASSIFICATION (CNN1D)

The advent of one-dimensional CNNs has eliminated the need for spectrogram transformations by enabling direct processing of raw time-series signals [37]. We carefully designed and iteratively refined a series of custom 1D CNN architectures, incorporating state-of-the-art techniques, such as skipping block and attention mechanism, to ultimately select the optimal model for processing time-domain ECG signals.

The initial network (v0) was adapted from LeNet [39], a compact architecture designed for digit classification. To accommodate time-series signals, we converted the 2D layers to 1D layers while preserving the original parameters. Additionally, we replaced the Flatten layer with a Global Average Pooling 1D layer. After developing the initial network (v0), we refined its architecture by adjusting layer parameters, such as the number of filters and kernel size (v1). We then investigated whether adding more layers could enhance performance (v2). Rather than simply increasing depth, we explored advanced techniques, including skip connections inspired by ResNet [40] (Fig. 6a) and a modified channel attention mechanism from EfficientNet [41] (Fig. 6b), to assess their impact (v3). Finally, we combined skip connections with an attention mechanism (Fig. 6c) and evaluated its performance (v4). For activation functions, we evaluated Swish (v5), GELU (v6), and ReLU (v7), with ReLU demonstrating superior performance.

Our Apnea-ECG model reached optimal performance with the v7 architecture (Fig. 7), whereas simpler designs sufficed for MIT-BIH and ECG-ID classification (Fig. 8). Section IV reveals that even this CNN1D architecture (without incorporating FAN technique) exceeds most literature results.

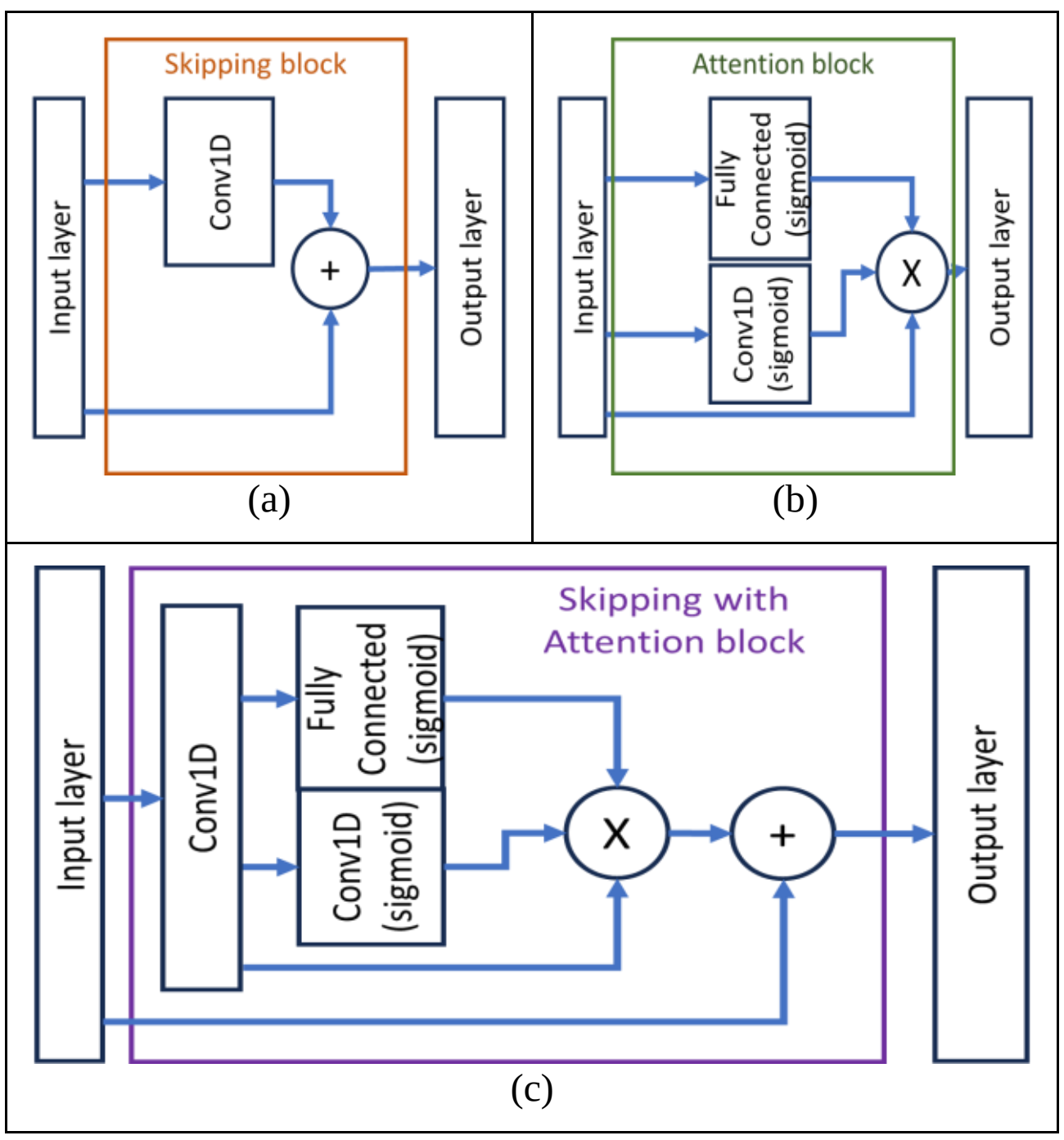

**FIGURE 6.** Blocks used in CNN1D and FFT1D models. (a) Skipping block. (b) Attention block. (c) Skipping with attention block.

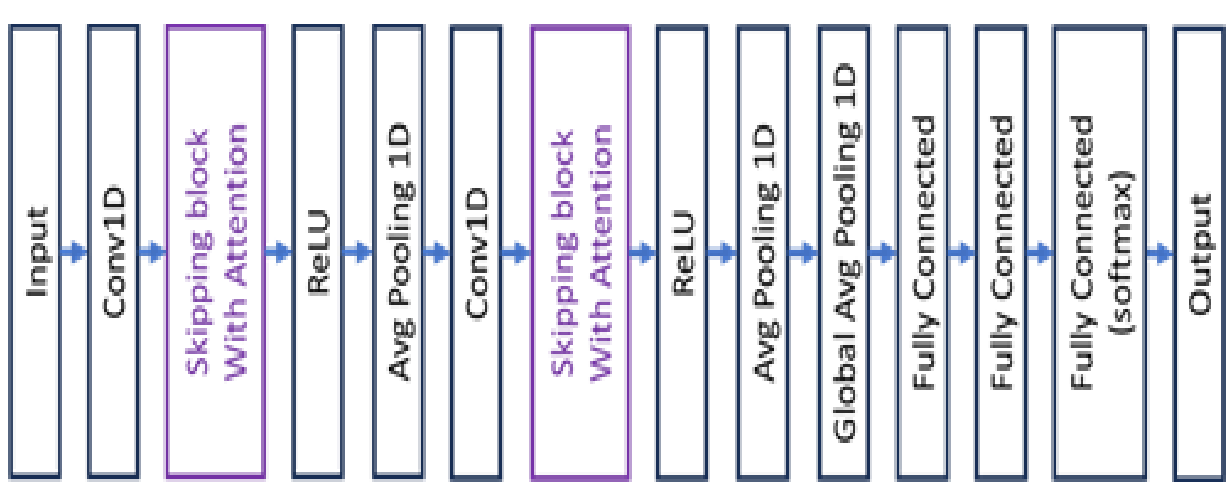

**FIGURE 7.** Architecture of the CNN1D and FFT1D models (v7) for Apnea-ECG problem.

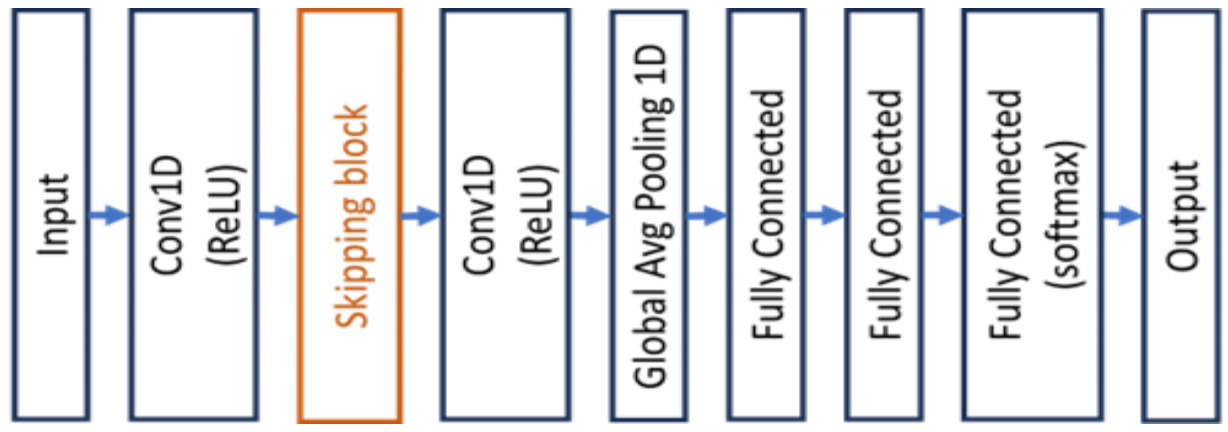

**FIGURE 8.** Architecture of the CNN1D and FFT1D models (v7) for MIT-BIH and ECG-ID.

### *D. 1-D FFT CLASSIFICATION (FFT1D)*

For frequency-domain classification, we transformed ECG segments via Fast Fourier Transform (FFT) [38]. We maintained the CNN1D (v7) architecture but used real/imaginary FFT components as inputs (Fig. 9). Although 1D-CNN's translation invariance is advantageous for time-domain analysis, this property does not extend to frequency-domain representations, which may explain its reduced effectiveness. Our experimental results in section IV confirm this observation, with FFT-based 1D-CNN (FFT1D) underperforming compared to the time-domain 1D-CNN (CNN1D).

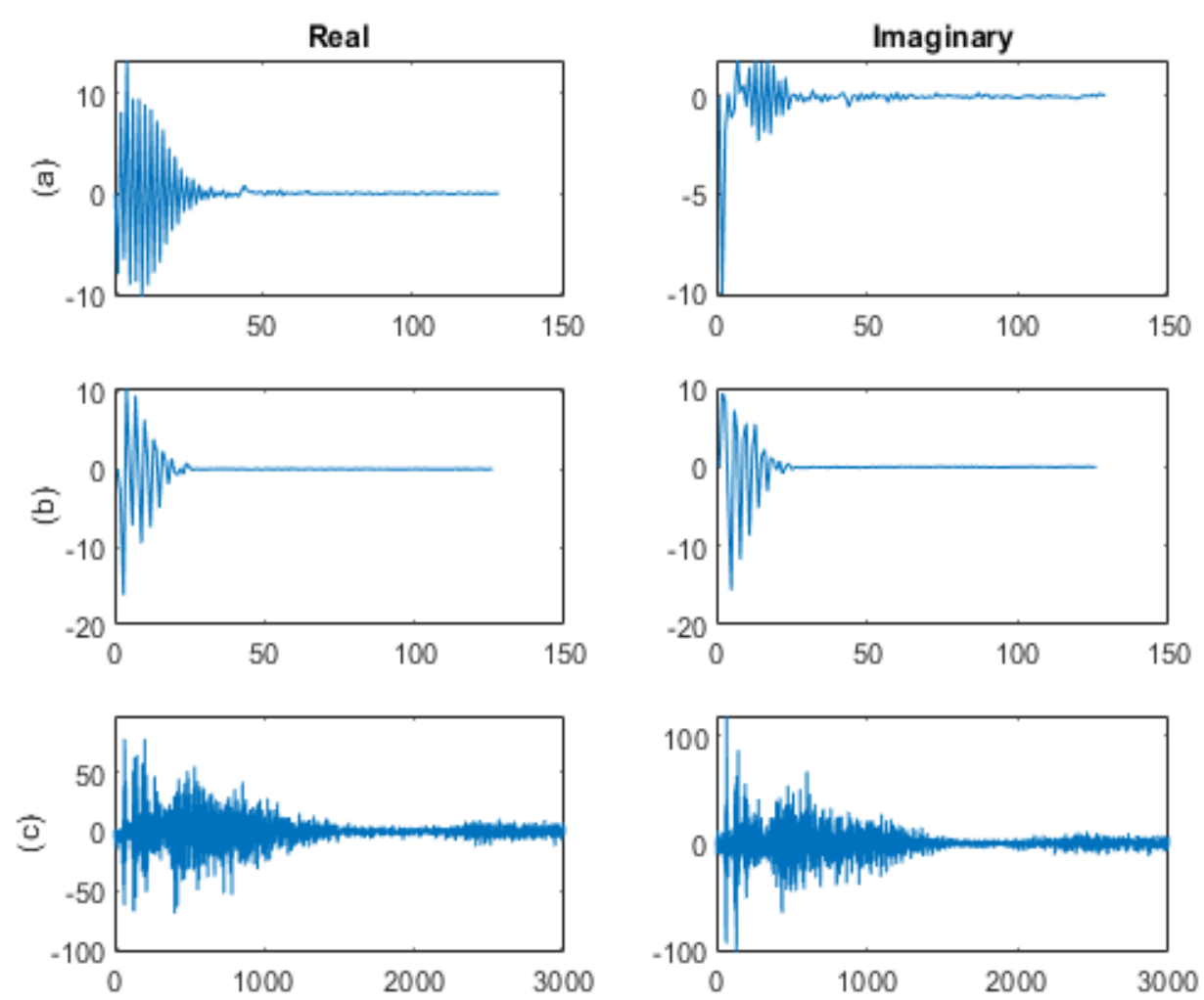

**FIGURE 9.** Real and imaginary components of the FFT-transformed ECG segments depicted in Fig. 3.

### *E. CNN1D-FAN*

To systematically evaluate the benefits of FAN integration, we modified our baseline CNN1D architecture (Figs. 7-8) by replacing all fully-connected (FC) layers — with the exception of the final softmax classification layer — with FC-FAN blocks with independent sine and cosine weights (Fig. 1b). This transformation preserves the network's fundamental one-dimensional temporal processing while introducing Fourier analysis capabilities through the FC-FAN components.

We maintained identical experimental conditions to ensure a fair comparison with the original CNN1D implementation. Despite these architectural enhancements, our quantitative analysis (section IV) revealed only marginal performance improvements.

### F. CONVOLUTIONAL FAN (CFAN)

To improve upon the CNN1D-FAN architecture's performance, we introduce the Convolutional FAN (CFAN), which extends Fourier-based processing by incorporating sine and cosine activations throughout the entire network — including both fully-connected (FC) and convolutional layers.

In convolutional neural networks (CNNs), the FC layers typically serve as the final classification stage rather than for feature extraction. Consequently, replacing only FC layers by FC-FAN blocks captures the periodic nature of the extracted features. To further embed periodic information throughout the network, we extended this concept to the convolutional layers.

We introduce the CONV-FAN block (Fig. 10b) to apply sine and cosine activation functions to Conv1D layers, enabling the network to better represent both the input signal and intermediate attributes with periodic characteristics.

We replaced all standard convolution layers in CNN1D-FAN architecture with CONV-FAN blocks. We also replaced the Conv1D layers in the Skipping blocks (Fig. 6a) with CONV-FAN, resulting in what we now designate as Skipping with CONV-FAN blocks. The Conv1D blocks in the attention block (Fig. 6b) were not replaced.

For the Apnea-ECG task, each CONV-FAN block consisted of three Conv1D layers, each containing 4 filters (totaling 12 filters) with a kernel size of 64 (Fig. 11). In the MIT-BIH and ECG-ID tasks, each CONV-FAN block comprised three Conv1D layers, each with 32 filters (totaling 96 filters) and a kernel size of 64 (Fig. 12).

In this architecture, we use separate weights for sine and cosine activations, both in FC-FAN and CONV-FAN blocks. The CFAN model, trained with identical parameters and using the same training, validation and testing sets as the original CNN1D-FAN, achieved superior accuracies across all datasets, detailed in section IV.

### G. CONVOLUTIONAL SHARED-MATRIX FAN (CSFAN)

The original FAN [34] employs shared weights for both *sine* and *cosine* activations. To investigate the impact of weight sharing, we replaced the FC-FAN and CONV-FAN blocks in the CFAN architecture (Figs. 11-12) with FC-SFAN and CONV-SFAN blocks. We refer to this variant architecture as CSFAN (Convolutional Shared-matrix FAN). The obtained results were inferior to CFAN (section IV)

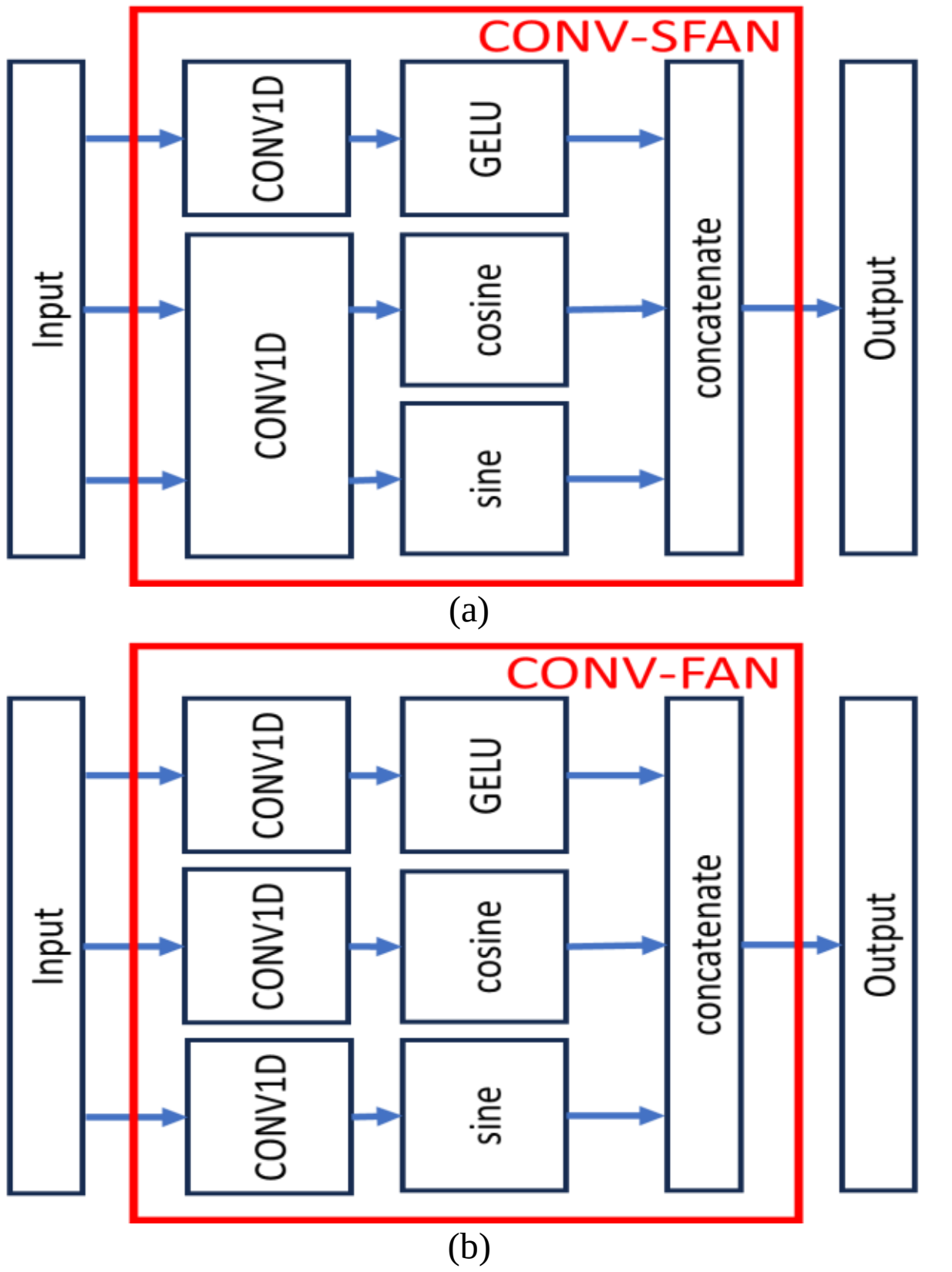

**FIGURE 10.** **(a) CONV-SFAN block uses weight-sharing between sine and cosine functions, with equal proportions (1:1) of GELU and trigonometric activations. (b) CONV-FAN block employs independent weights for sine and cosine, implementing a 1:1:1 ratio of GELU, sine, and cosine activation paths).**

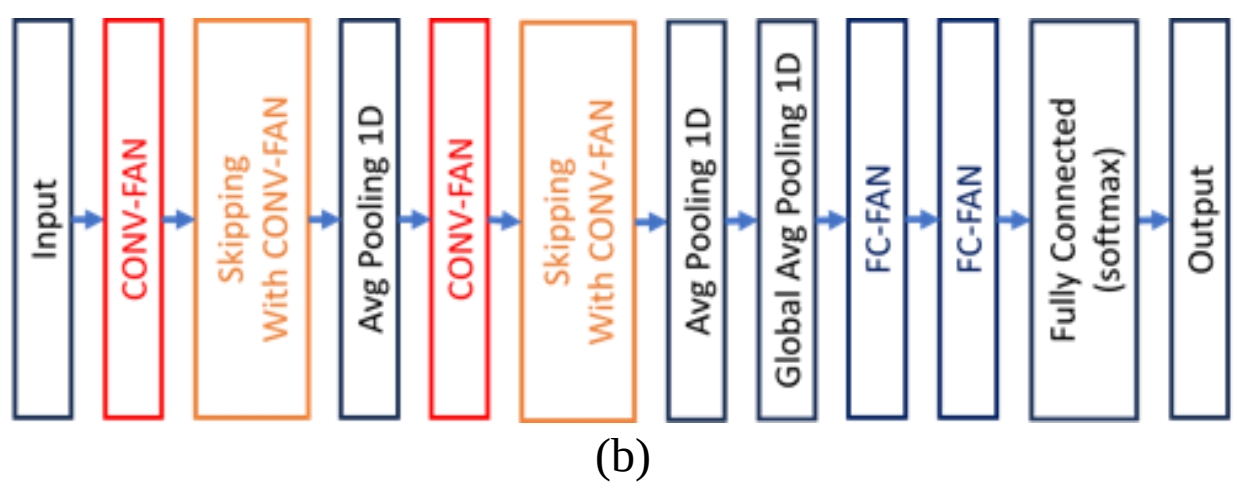

**FIGURE 11.** **CFAN architecture for Apnea-ECG problem).**

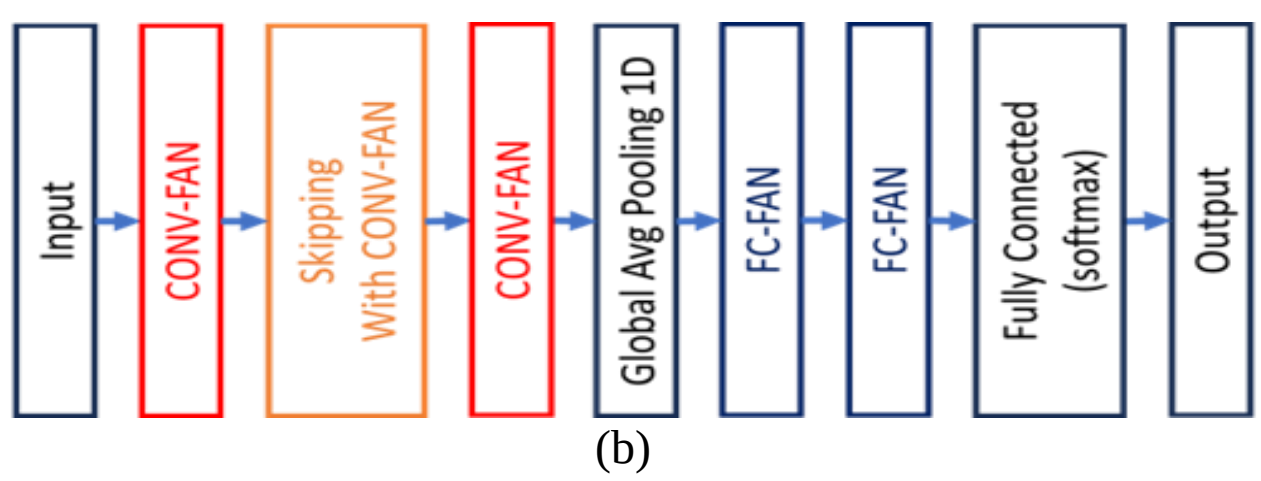

**FIGURE 12.** **CFAN architecture for MIT-BIH and ECG-ID tasks.**

### H. EVALUATION METRICS

To evaluate performance on the binary classification task (Apnea-ECG), we computed the ROC-AUC (Receiver Operating Characteristic – Area Under the Curve) and the accuracy for each test fold. We also computed Precision, Recall and F1-score considering threshold of 0.5.

Accuracy for multi-class problems (MIT-BIH with N = 5 classes and ECG-ID with N = 90 classes) was computed by counting the number of correctly classified samples — those for which the class with the highest probability matched the true class label — and dividing by the total number of samples. To compute Precision and Recall, we approached the multi-class problems as a series of binary classification tasks, treating each class in turn as the positive class and the rest as negative (using threshold of 0.5), resulting in N values per metric. The final score for each fold was obtained by averaging these N values. The average Precision and Recall were then used to calculate the F1-score as follows:

$$F1\text{-}score = \frac{2*AvPrecision*AvRecall}{AvPrecision+AvRecall}. \quad (7)$$

The AUC for multi-class problems was also computed considering the multi-class problems as a series of binary classification problems.

We note that Precision, Recall, and F1-score are threshold-dependent metrics (with a threshold of 0.5 in our case) and thus have inherent limitations. Although AUC offers a threshold-independent evaluation, it tends to saturate near maximum in our tasks. Therefore, we primarily use accuracy as our preferred metric for method comparison.

## IV. RESULTS

### A. IMPLEMENTATION

We implemented the techniques described in this paper using TensorFlow v2.10.1 and Keras v2.10.0. All experiments were conducted on a machine running Windows 11, equipped with an AMD Ryzen 7 5800X processor, 64 GB of RAM, and an NVIDIA GeForce RTX 3060 GPU with 12 GB of VRAM. Table I summarizes the typical training and testing times per fold, along with the number of trainable parameters.

TABLE I
TRAINING / EVALUATION TIMES IN SECONDS FOR ONE-FOLD AND THE NUMBER OF TRAINABLE PARAMETERS OF THE ARCHITECTURE.

| | MIT-BIH | ECG-ID | Apnea-ECG |
|---|---|---|---|
| SPECT | 8712.3 / 15.52<br>20,439,389 | 1872.0 / 2.26<br>20,446,614 | 1017.8 / 2.41<br>20,439,134 |
| CNN1D | 503.6 / 0.20<br>1,208,309 | 166.2 / 0.21<br>1,215,534 | 1562.5 / 0.37<br>40,982 |
| FFT1D | 415.8 / 0.39<br>1,208,309 | 219.1 / 0.21<br>1,215,534 | 694.1 / 0.32<br>40,982 |
| FAN | 385.9 / 0.44<br>1,208,309 | 170.7 / 0.21<br>1,215,534 | 1535.2 / 0.38<br>40,982 |
| CFAN | 655.0 / 0.60<br>1,208,309 | 235.9 / 0.22<br>1,215,534 | 1837.9 / 0.75<br>40,982 |
| CSFAN<br>(CFAN variation) | 534.6 / 0.50<br>809,315 | 165.1 / 0.21<br>816,540 | 1458.3 / 0.56<br>29,540 |

### B. PERFORMANCE RESULTS

Tables II and III present the comprehensive performance evaluation across all models.

TABLE II
MEAN ± SD VALUES FOR AUC (TOP) AND ACCURACY (%, BOTTOM).

| | MIT-BIH | ECG-ID | Apnea-ECG |
|---|---|---|---|
| SPECT | 0.9948±0.0022<br>98.93±0.50 | **0.9999±0.0001**<br>95.34±0.39 | 0.9376±0.0679<br>89.86±2.74 |
| CNN1D | 0.9955±0.0025<br>98.92±0.15 | 0.9991±0.0012<br>95.20±1.14 | 0.9847±0.0036<br>93.90±1.07 |
| FFT1D | 0.9941±0.0029<br>98.66±0.24 | 0.9990±0.0008<br>94.06±0.31 | 0.9439±0.0140<br>87.55±1.92 |
| FAN | 0.9952±0.0034<br>98.94±0.09 | 0.9997±0.0002<br>95.28±1.01 | 0.9833±0.0033<br>93.94±0.72 |
| CFAN | **0.9966±0.0024**<br>**98.95±0.17** | 0.9998±0.0002<br>**96.83±0.72** | **0.9877±0.0015**<br>**95.01±0.44** |
| CSFAN<br>(CFAN<br>variation) | 0.9958±0.0024<br>98.83±0.10 | **0.9999±0.0001**<br>95.93±1.01 | 0.9873±0.0018<br>94.30±0.79 |

TABLE III
CLASSIFICATION PERFORMANCE (MEAN ± SD %): PRECISION (TOP), RECALL (MIDDLE), F1-SCORE (BOTTOM) AT THRESHOLD=0.5. SEE SECTION III-H FOR DETAILS.

| | MIT-BIH | ECG-ID | Apnea-ECG |
|---|---|---|---|
| SPECT | 95.50±1.25<br>91.27±3.92<br>93.07±2.72 | 95.20±0.98<br>94.16±0.68<br>93.78±0.58 | 89.64±2.23<br>88.78±3.73<br>89.04±3.27 |
| CNN1D | **95.85±1.23**<br>91.72±1.44<br>93.57±1.07 | 95.13±1.67<br>93.82±1.59<br>93.61±1.87 | 93.52±1.08<br>93.55±1.33<br>93.51±1.17 |
| FFT1D | 95.34±1.36<br>90.07±3.26<br>92.33±2.30 | 94.20±1.10<br>92.80±0.48<br>92.46±0.44 | 87.01±2.08<br>86.40±2.11<br>86.64±2.07 |
| FAN | 95.28±0.55<br>92.09±1.50<br>93.55±0.91 | 94.91±2.37<br>93.73±1.31<br>93.45±1.72 | 93.54±0.83<br>93.60±0.82<br>93.55±0.77 |
| CFAN | 95.54±1.21<br>**92.88±1.21**<br>**94.12±0.98** | **97.10±0.92**<br>**95.84±0.60**<br>**95.77±0.72** | **94.77±0.47**<br>**94.63±0.72**<br>**94.67±0.49** |
| CSFAN<br>(CFAN variation) | 95.53±1.01<br>91.87±1.34<br>93.54±0.78 | 96.16±0.96<br>95.09±1.15<br>94.95±1.29 | 93.96±0.94<br>93.95±0.80<br>93.94±0.82 |

The accuracy results establish a performance hierarchy, from lowest to highest (Table II): FFT1D, followed by SPECT, then CNN1D and CNN1D-FAN with nearly tied performance, and finally CFAN which consistently achieved the highest accuracy. This ranking pattern remains the same when examining F1-scores (Table III). The minimal observed differences in AUC values preclude meaningful comparative interpretation of this metric.

CSFAN is a variant of CFAN that uses a shared weight matrix for the sine and cosine activations. Table II shows that CFAN outperforms CSFAN. Thus, it is preferable to use CFAN with independent weights rather than CSFAN with shared weights.

To rigorously assess CFAN's performance advantage, we conducted a paired Student's t-test comparing its classification accuracy against the next best-performing architecture (CNN1D-FAN) across two datasets: ECG-ID and Apnea-ECG. For the MIT-BIH dataset, we did not perform significance testing due to a ceiling effect, as accuracy approached 99% with minimal variance.

As shown in Table IV, CFAN outperforms CNN1D-FAN with statistical significance (p=0.02 for ECG-ID and p=0.0009 for Apnea-ECG). Since CNN1D-FAN already outperforms CNN1D, SPECT, and FFT1D, we can conclude that CFAN is statistically superior to all these methods.

TABLE IV
STATISTICAL SIGNIFICANCE (*P*-VALUES) OF ACCURACY DIFFERENCES (PAIRED *T*-TESTS): CFAN VS. CNN1D-FAN (SECOND-BEST PERFORMING MODEL).

| | ECG-ID | Apnea-ECG |
|---|---|---|
| CFAN vs CNN1D-FAN | 0.02 | 0.0009 |

Note that the CNN1D, FFT1D, CNN1D-FAN, and CFAN architectures were designed with identical numbers of trainable parameters, enabling direct performance comparisons and fair architectural evaluation. In these controlled comparisons, CFAN consistently demonstrated superior performance. CSFAN, on the other hand, was designed to maintain the same number of filters as CFAN.

### C. COMPARISON WITH LITERATURE

Table V compares CFAN's classification accuracy against prior work for each task, while Table VI provides additional performance comparisons using precision, recall, and F1-score metrics. We note that direct comparisons should be interpreted cautiously due to methodological variations across different studies.

TABLE V
COMPARISON OF CFAN'S ACCURACY WITH PUBLISHED RESULTS IN THE LITERATURE. THE RESULTS FOR APNEA-ECG DATASET ARE NOT DIRECTLY COMPARABLE, AS EACH STUDY EMPLOYS A DIFFERENT TESTING METHODOLOGY.

| Database | Authors | Year | Accuracy (%) |
|---|---|---|---|
| MIT-BIH | Sharma *et al.* [9] | 2024 | 98.67 |
| | Nemissi *et al.* [13] | 2024 | 98.78 |
| | ***CFAN*** | **2025** | **98.95** |
| ECG-ID | Nemirko and Lugovaya [3] | 2005 | 96.00 |
| | Fatima *et al.* [5] | 2022 | 93.41 |
| | Yi *et al.* [18] | 2024 | 94.26 |
| | **CFAN** | **2025** | **96.83** |
| Apnea-ECG | Zhou *et al.* [11] | 2022 | 90.93 |
| | Misra *et al.* [26] | 2022 | 93.01 |
| | Bhongade *et al.* [19] | 2023 | 94.77 |
| | **Li *et al.* [25]** | **2023** | **98.31** |
| | Zhao *et al.* [22] | 2024 | 91.02 |
| | Nguyen *et al.* [23] | 2024 | 92.11 |
| | Gupta *et al.* [27] | 2024 | 96.60 |
| | CFAN | 2025 | 95.01 |

TABLE VI
PERFORMANCE COMPARISON (MEAN ± SD %) OF PRECISION (TOP), RECALL (MIDDLE), AND F1-SCORE (BOTTOM) WITH PUBLISHED LITERATURE. DIRECT CROSS-STUDY COMPARISONS ARE LIMITED DUE TO METHODOLOGICAL DIFFERENCES. SEE SECTION IV-C FOR DETAILS.

| Authors | MIT-BIH | ECG-ID | Apnea-ECG |
|---|---|---|---|
| Nemissi *et al.* [13] | 92.61<br>94.61<br>93.85 | - | - |
| Sharma *et al.* [9] | **97.80**<br>**96.00**<br>**96.80** | - | - |
| Li *et al.* [25] | - | - | **96.45**<br>**97.66**<br>**97.05** |
| Zhao *et al.* [22] | - | - | -<br>84.25<br>87.56 |
| Nguyen *et al.* [23] | - | - | 91.3<br>87.7<br>89.5 |
| Misra *et al.* [26] | - | - | 93.15<br>92.93<br>93.04 |
| Bhongade *et al.* [19] | - | - | 93.80<br>92.55<br>93.07 |
| CFAN | 95.54±1.21<br>92.88±1.21<br>94.12±0.98 | **97.10±0.92**<br>**95.84±0.60**<br>**95.77±0.72** | 94.77±0.47<br>94.63±0.72<br>94.67±0.49 |

Our segmentation methodology for the MIT-BIH database adheres to standard conventions. CFAN achieves state-of-the-art classification accuracy (98.95%, Table V), outperforming the previous best reported result of 98.78%.

The method proposed by Sharma *et al*. [9] outperforms CFAN in terms of precision, recall, and F1-score (Table VI), although CFAN achieves higher accuracy (Table V). This discrepancy is likely due to their simplified binary evaluation framework (reference class vs. normal rhythms), as opposed to the comprehensive multiclass evaluation used in our study. Their approach represents a fundamentally less complex classification task, highlighting the importance of consistent and comparable evaluation methodologies when assessing model performance.

While Nemissi *et al*. [13] achieve marginally higher recall than CFAN (Table VI), CFAN demonstrates superior performance across all other metrics (accuracy, precision, and F1-score). Note that recall alone offers limited insight as it exclusively measures true positive rates.

For the ECG-ID database, there is some variation in the method used for segment extraction. In our work, we employed the same method as Nemirko and Lugovaya [3], which is similar to the approach used by Yi *et al*. [18], but differs from the method used by Zhao *et al*. [16]. Compared to comparable studies, CFAN achieved the highest accuracy, with a result of 96.83%, whereas the best previous accuracy was 96.00% (Table V). For identity recognition, accuracy is the primary evaluation metric, and other metrics are typically not reported. As a result, there are no comparative values available from other studies.

For the Apnea-ECG database, significant variations in the number and choice of samples used for training and testing across studies make it difficult to directly compare our results with those in the literature. See the brief description of the differences in section II-E.

## V. CONCLUSIONS

This study presented the Convolutional Fourier Analysis Network (CFAN), a novel deep learning architecture that effectively integrates time- and frequency-domain analysis for ECG classification. Through comprehensive experimentation across three clinically relevant tasks — arrhythmia classification (MIT-BIH), identity recognition (ECG-ID), and sleep apnea detection (Apnea-ECG) — CFAN demonstrated superior performance compared to conventional approaches, including spectrogram-based 2D CNNs (SPECT), time-domain 1D CNNs (CNN1D), frequency-domain 1D CNNs (FFT1D), and the Fourier Analysis Network (CNN1D-FAN).

Key contributions of this work include:

*Architectural innovation*: CFAN's CONV-FAN blocks, which incorporate sine and cosine activations in convolutional layers with independent weight matrices for sine and cosine, enable joint time-frequency feature learning without spectrogram conversion.

*State-of-the-Art performance*: CFAN achieved benchmark accuracy (98.95% MIT-BIH, 96.83% ECG-ID, 95.01% Apnea-ECG) and statistically significant improvements over the second-best CNN1D-FAN ($p=0.02$ in ECG-ID and $p=0.0009$ in Apnea-ECG).

*Theoretical insight*: The success of CFAN validates that explicit modeling of periodicity — via Fourier-inspired activations — enhances feature extraction for biomedical signals.

In summary, CFAN advances ECG analysis by unifying time and frequency representations in an end-to-end trainable framework. Its consistent outperformance of existing methods, coupled with theoretical grounding in Fourier principles, positions it as a versatile tool for biomedical signal processing.

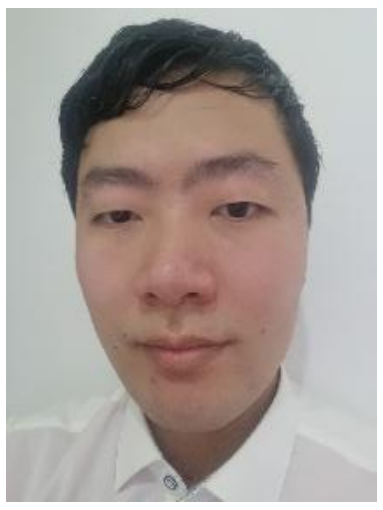

**SAM JEONG** received the B.S. degree in electrical engineering with emphasis in electronic systems in 2017, followed by M.S. degree in electrical engineering with area of concentration in biomedical engineering in 2021, both from the University of São Paulo (USP), Brazil. He is currently pursuing the Ph.D. degree in electrical engineering with the Department of Electronic Systems, USP.

His research interest includes signal and image processing, medical signal and image analysis, and machine learning.

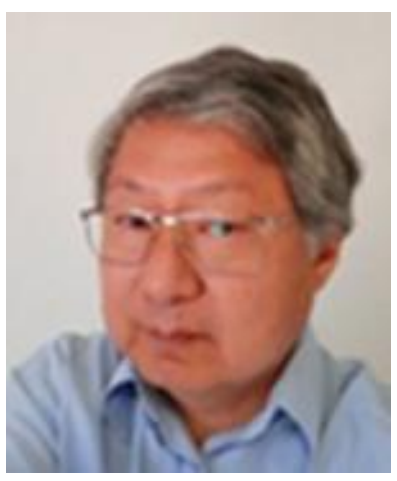

**HAE YONG KIM** received his B.S. and M.S. degrees (Hons.) in computer science and his Ph.D. in electrical engineering from the University of São Paulo (USP), Brazil, in 1988, 1992, and 1997, respectively.

He is currently an Associate Professor in the Department of Electronic Systems Engineering at USP. He has authored over 100 articles and holds three patents. His research interests include image processing, machine learning, medical image analysis, and computer security.

Dr. Kim and his colleagues have received several prestigious awards, including the Petrobras Technology Award (6th edition) in the Refining and Petrochemical Technology category (2013), the Best Paper in Image Analysis Award at the Pacific-Rim Symposium on Image and Video Technology (2007), and the Thomson ISI Essential Science Indicators “Hot Paper” Award (2005) for publishing one of the top 0.1% most cited papers in computer Science.